\documentclass[a4paper,twocolumn,fleqn]{article}
\usepackage{ist}
\usepackage{times}
\usepackage{epsfig}
\usepackage{graphicx}
\usepackage{amsmath}
\usepackage{amssymb}
\usepackage{graphicx}
\usepackage{multirow}
\usepackage{color}
\usepackage{cite}
\usepackage{dsfont}
\usepackage[table,xcdraw]{xcolor}
\usepackage{ragged2e}
\usepackage{nccmath}
\usepackage{textcomp}
\usepackage{arydshln}
\usepackage{float}
\usepackage{hyperref}

\DeclareMathOperator*{\argmin}{arg\,min}
\title{Interactive White Balancing for Camera-Rendered Images}
\author{Mahmoud~Afifi and Michael~S.~Brown; York University, Toronto}
\date{}
\begin{document}
\maketitle
\thispagestyle{empty}
\begin{abstract}\vspace{1mm}
White balance (WB) is one of the first photo-finishing steps used to render a captured image to its final output. WB is applied to remove the color cast caused by the scene's illumination. Interactive photo-editing software allows users to manually select different regions in a photo as examples of the illumination for WB correction (e.g., clicking on achromatic objects).  Such interactive editing is possible only with images saved in a RAW image format.  This is because RAW images have no photo-rendering operations applied and photo-editing software is able to apply WB and other photo-finishing procedures to render the final image. Interactively editing WB in camera-rendered images is significantly more challenging.  This is because the camera hardware has already applied WB to the image and subsequent nonlinear photo-processing routines.  These nonlinear rendering operations make it difficult to change the WB post-capture. The goal of this paper is to allow interactive WB manipulation of camera-rendered images.   The proposed method is an extension of our recent work~\cite{afifi2019color} that proposed a post-capture method for WB correction based on nonlinear color-mapping functions.  Here, we introduce a new framework that links the nonlinear color-mapping functions directly to user-selected colors to enable {\it interactive} WB manipulation. This new framework is also more efficient in terms of memory and run-time (99\% reduction in memory and 3$\times$ speed-up).  Lastly, we describe how our framework can leverage a simple illumination estimation method (i.e., gray-world) to perform auto-WB correction that is on a par with the WB correction results in~\cite{afifi2019color}. The source code is publicly available at \href{https://github.com/mahmoudnafifi/Interactive_WB_correction}{https://github.com/mahmoudnafifi/Interactive\_WB\_correction}.

\end{abstract}

\section{1. Introduction}\label{sec:introduction} \vspace{1mm}

A number of photo-finishing steps are applied onboard a camera to render the RAW sensor image to the final output images.  One of the earliest and most critical steps in this rendering process is the white-balance (WB) correction that is applied to remove color casts caused by the scene illumination.  The challenge for performing WB correctly is estimating the scene's illumination from a captured RAW image.  Illumination estimation is performed by the camera's auto-white-balance (AWB) module that estimates a 3D color vector representing the color cast caused by the illumination.  The WB correction itself is a simple diagonal matrix operation applied to scale each RGB color channel based on the estimated illumination vector.  If WB is applied incorrectly, or based on an estimated illumination that the user does not prefer, the image will have a noticeable color cast that is difficult to adjust post-capture.  This difficulty arises because several nonlinear photo-finishing operations are applied on the camera after the WB operation.

\begin{figure}[t]
\centering
\includegraphics[width=0.95\linewidth]{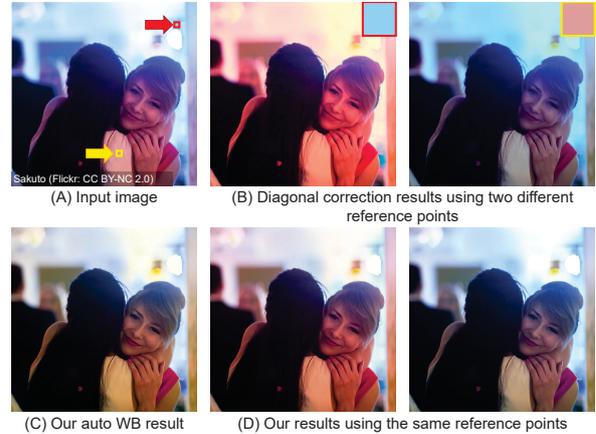}
\caption{(A) A camera-rendered image with the wrong WB applied.  The user selects two reference colors in the scene representing achromatic scene materials. (B) The results correction using the conventional diagonal WB correction matrix.  Due to the nonlinear camera-rendering, the conventional approach is not sufficient to correct the WB.  (C) Our method's AWB on (A).  (D) Our method's results using the user-supplied reference colors.\vspace{-4mm}}
\label{fig:teaser}

\end{figure}

There is photo-editing software (e.g., Adobe Lightroom~\cite{kelby2014adobe},  Skylum~\cite{Skylum}, Affinity Photo~\cite{affinity}) that enables interactive WB manipulation.  Instead of applying AWB, these methods allow the user to select a pixel's RGB values in the image of achromatic scene materials to serve as the estimated illumination color vector.  In some scenes, there may be more than one illuminant present and the users can choose which illumination they prefer to correct (see~Fig.~\ref{fig:teaser}).  This interactive WB editing, however, is possible only for photos saved in a RAW image format.  This is because RAW images have no photo-finishing applied---instead, the photo-editing software mimics the onboard camera rendering using the user-supplied parameters.

The goal of this paper is to allow interactive WB manipulation for camera-rendered images.  As previously mentioned, WB manipulation in camera-rendered images is challenging due to the nonlinear operations applied by the camera hardware.  In recent work~\cite{afifi2019color}, we showed that even when an exact achromatic reference scene point is known in a camera-rendered image, the conventional diagonal correction method cannot sufficiently remove the color casts caused by WB errors (see~Fig.~\ref{fig:teaser}). To address this issue,~\cite{afifi2019color} proposed an effective nonlinear polynomial color correction function in lieu of the conventional diagonal WB matrix.  The work in ~\cite{afifi2019color} used a histogram feature computed from an input image to determine which nonlinear correction function to use to correct a camera-rendered image that had the wrong WB applied.  The work in ~\cite{afifi2019color}, however, provided no mechanism to link the nonlinear WB color-mapping functions to the user's selected pixel values.

\vspace{-2mm}
\paragraph{Contribution}~
We build upon the idea in ~\cite{afifi2019color} and propose a new framework that allows interactive WB editing in camera-rendered images. Our approach works by associating color-cast vectors with rectification functions that output nonlinear color-mapping functions to correct the camera-rendered image's WB.  This allows the user to supply a color vector from the image that results in a nonlinear color mapping to modify the image's WB based on the specified color vector.  Our framework requires only 1\% of the memory used in ~\cite{afifi2019color} and runs 3$\times$ faster, allowing interactive functionality (see~Fig.~\ref{fig:teaser}).  In addition, we show that our framework can be used to perform AWB correction for camera-rendered images that were incorrectly white-balanced with results on a par with ~\cite{afifi2019color}.

\section{2. Related Work}\label{sec:relatedwork}  \vspace{1mm}

 \begin{figure*}[t]
   \centering
   \includegraphics[width=0.98\linewidth]{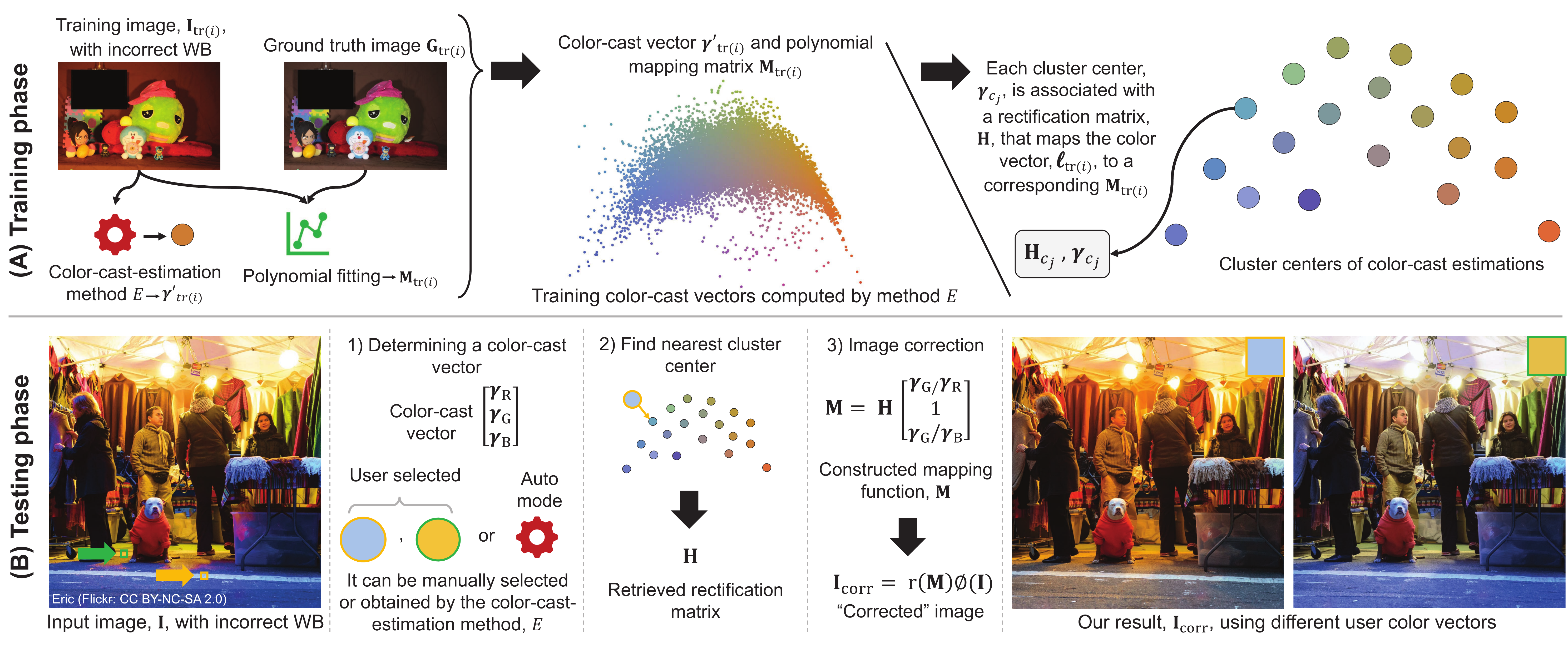}
   \caption{Overview of our proposed method. (A) [Training Phase] For each image in the training dataset, we estimate a color-cast vector -- shown here projected into the $rg$ chromaticity space to aid visualization -- using an off-the-shelf-illumination estimation algorithm $E$ (e.g., gray-world \cite{buchsbaum1980spatial}). A nonlinear color-mapping function that maps this training image to a correctly white-balanced image is also computed. The color-cast vectors of each training image are clustered.  A \textit{rectification function} is computed for each cluster that returns a color-mapping function based on a color-cast vector.
   (B) [Testing Phase] When applying our method, the user either manually provides a color-cast vector or uses the method $E$ to predict a color-cast vector. Using this color-cast vector, the most similar cluster in the training data is found and its rectification function is used to compute a mapping function $\mathbf{M}$ that is applied to correct the image.\vspace{-4mm}}
   \label{fig:main}
   \end{figure*}

We first discuss related work targeting auto WB correction in Sec. 2.1, followed by work related to user-interactive WB correction approaches in Sec. 2.2.

\subsection{2.1. Auto White-Balance Correction}\label{sec:auto-wb}  \vspace{1mm}
Cameras perform WB correction to remove color casts caused by scene illumination. This procedure is applied to the linear scene-referred images (i.e., RAW sensor images) and is either performed in a fully automated mode or based on a manually selected WB pre-set (e.g., Daylight, Shade, Tungsten, Fluorescent). In the AWB mode, cameras employ illuminant estimation algorithms to estimate the scene illumination color; representative examples of illuminant estimation methods include \cite{buchsbaum1980spatial, cardei1999committee, GE, cardei2002estimating, xiong2006estimating, SoG, gehler2008bayesian, finlayson2013corrected, gijsenij2012improving, bianco2015color, cheng2015effective,  barron2015convolutional, shi2016deep, hu2017fc, barron2017fast, qian2019finding, bianco2019quasi, Afifi2019Sensor, hernandez2020multi}. After determining the scene illuminant color, a linear diagonal correction matrix is used to remove the effect of the scene illuminant color~\cite{gijsenij2011computational}. Afterwards, cameras apply a set of operations to render the RAW values into the final display-referred colors represented in one of the standard display-referred color spaces---for example, the standard RGB (sRGB) space \cite{anderson1996proposal}.

Once the image is rendered it is difficult to modify the image's WB to either correct an improperly applied WB or adjust the image for aesthetic purposes~\cite{afifi2019color, afifi20deepWB}. This difficulty is due to the nonlinear camera-specific rendering procedures applied after the WB correction. As a consequence, utilizing such linear correction for camera-rendered image white balancing mostly fails to produce satisfactory results, as discussed in \cite{afifi2019color}.

Recently, we showed in~\cite{afifi2019color} that correcting WB for camera-rendered images can be achieved using nonlinear color-mapping functions or via the nonlinearity provided in deep learning (DL) networks \cite{afifi20deepWB}. Our work is inspired by the effectiveness of these nonlinear color-mapping functions to adjust WB in camera-rendered images. However, in contrast to \cite{afifi2019color}, our method provides a mechanism to link these mapping functions to color vectors found in the images.  This enables the interactive editing ability that is the focus of this paper.

\subsection{2.2. User-Interactive White-Balance Correction}\label{seec:inter-wb}  \vspace{1mm}

Prior work dedicated to post-capture interactive WB correction does allow interactive WB modification, but only in the form of a color temperature slider \cite{afifi20deepWB, afifi2019colortemp}.  For example, Afifi et al. \cite{afifi2019colortemp} proposed embedding metadata in camera-rendered images at capture time that allowed for post-capture WB modification. In their method, a color temperature slider is used to interpolate between different pre-defined color temperatures. Later, Afifi and Brown \cite{afifi20deepWB} introduced a deep-learning (DL) framework for WB correction and modification. The DL framework ``re-renders'' the input image with two different indoor/outdoor WB settings and allows the user to manipulate a temperature slider to interpolate between the two indoor/outdoor WB images.

Abdelhamed et al. \cite{abdelhamed2019markwhite} showed through a set of user studies that the color temperature slider is less preferred compared to directly marking achromatic reference points present in the scene. This approach allows the user to mark a known achromatic reference point that can represent current scene illumination color and was used directly by the camera's hardware to render the image. Such manual selection cannot properly work when the image is rendered with strong color casts caused by camera WB errors, as demonstrated in \cite{afifi2019color}. In contrast to existing approaches for post-capture user-interactive WB correction, our method allows the user to select any achromatic reference point to change the WB setting of the camera-rendered image.

\section{3. Methodology}\label{sec:method}  \vspace{1mm}

Given an input image, $\mathbf{I}$, that is rendered with an incorrect or undesired WB setting, the goal is to generate a new output image, $\mathbf{I}_{\texttt{corr}}$, that represents the input $\mathbf{I}$ as it would appear if re-rendered with a new (presumably correct or desired) WB setting.  We will refer to the target  ``ground truth'' white-balanced image as $\mathbf{G}$. In the remaining part of this paper, each image is represented as $3\!\times\!N$ matrix of the $\texttt{R}$, $\texttt{G}$, $\texttt{B}$ triplets, where $N$ is the total number of pixels in the image.

A traditional diagonal-based solution relies on determining a 3D vector $\pmb{\gamma} = \left[\pmb{\gamma}(\texttt{R}), \pmb{\gamma}({\texttt{G}}), \pmb{\gamma}({\texttt{B})}\right]^\top$ that represents the scene illuminant color.  Since in our application, we are applying a correction not of the scene illumination but instead to a color cast present in the camera-rendered image with the wrong WB applied, we will refer to $\pmb{\gamma}$ as a color-cast vector instead of an illumination vector.  In our case, this color-cast vector is provided manually by the user or based on an algorithm when in AWB mode. Given a color-cast vector, the image is assumed to be corrected using the following equation:
\begin{equation}
\label{diagonal_corr}
\mathbf{I}_{\texttt{corr(diag)}} = diag\left(\pmb{\ell}\right)\textrm{ }\mathbf{I},
\end{equation}
\noindent
where $diag\left(\cdot\right)$ is a $3\!\times\!3$ diagonal matrix of the {\it color-cast-correction} vector, $\pmb{\ell}$. The vector $\pmb{\ell} = \left[\pmb{\gamma}(\texttt{G})/\pmb{\gamma}(\texttt{R}), \texttt{ } \texttt{1}, \texttt{ } \pmb{\gamma}(\texttt{G})/\pmb{\gamma}(\texttt{B})\right]^\top$ and represents a simple modification of the color-cast vector.

Due to the nonlinearity applied to camera-rendered images, this simple scaling operation cannot properly correct WB errors. In ~\cite{afifi2019color}, the diagonal correction matrix was replaced with a polynomial color-mapping function that could deal with the nonlinearities in the input $\mathbf{I}$. This function was computed in the following form:
\begin{equation}
\label{eq:polynomial}
\mathbf{I}_{\texttt{corr(poly)}} = r\left(\mathbf{M}\right)\textrm{ }\phi\left(\mathbf{I}\right),
\end{equation}
\noindent
where $\phi$ is a kernel mapping function \cite{hong2001study},\ s.t.\ $\phi:\phi\left([\texttt{R}, \texttt{ }\texttt{G}, \texttt{ }\texttt{B}]^\top\right) \rightarrow [\texttt{R}$, $\texttt{G}$, $\texttt{B}$, $\texttt{RG}$, $\texttt{RB}$, $\texttt{GB}$, $\texttt{R}^2$, $\texttt{G}^2$, $\texttt{B}^2$, $\texttt{RGB}$, $1]^\top$, $\mathbf{M} \in \mathbb{R}^{33}$ is a vectorized form of the polynomial mapping matrix, and $r\left(\cdot\right)$ is a reshaping function that constructs the $3\!\times\!11$ matrix from the vectorized version, $\mathbf{M}$. This polynomial matrix can be computed in a closed-form via standard least squares methods.

Though this nonlinear color mapping achieves better results compared to the diagonal-based correction, it lacks a correlation to a color-cast vector. In the following section, we describe how to efficiently associate these color-mapping functions with color vectors that can be used for WB correction in a camera-rendered image $\mathbf{I}$.   Fig.~\ref{fig:main} provides an overview of our framework.

\begin{figure}[t]
 \centering
 \includegraphics[width=0.95\linewidth]{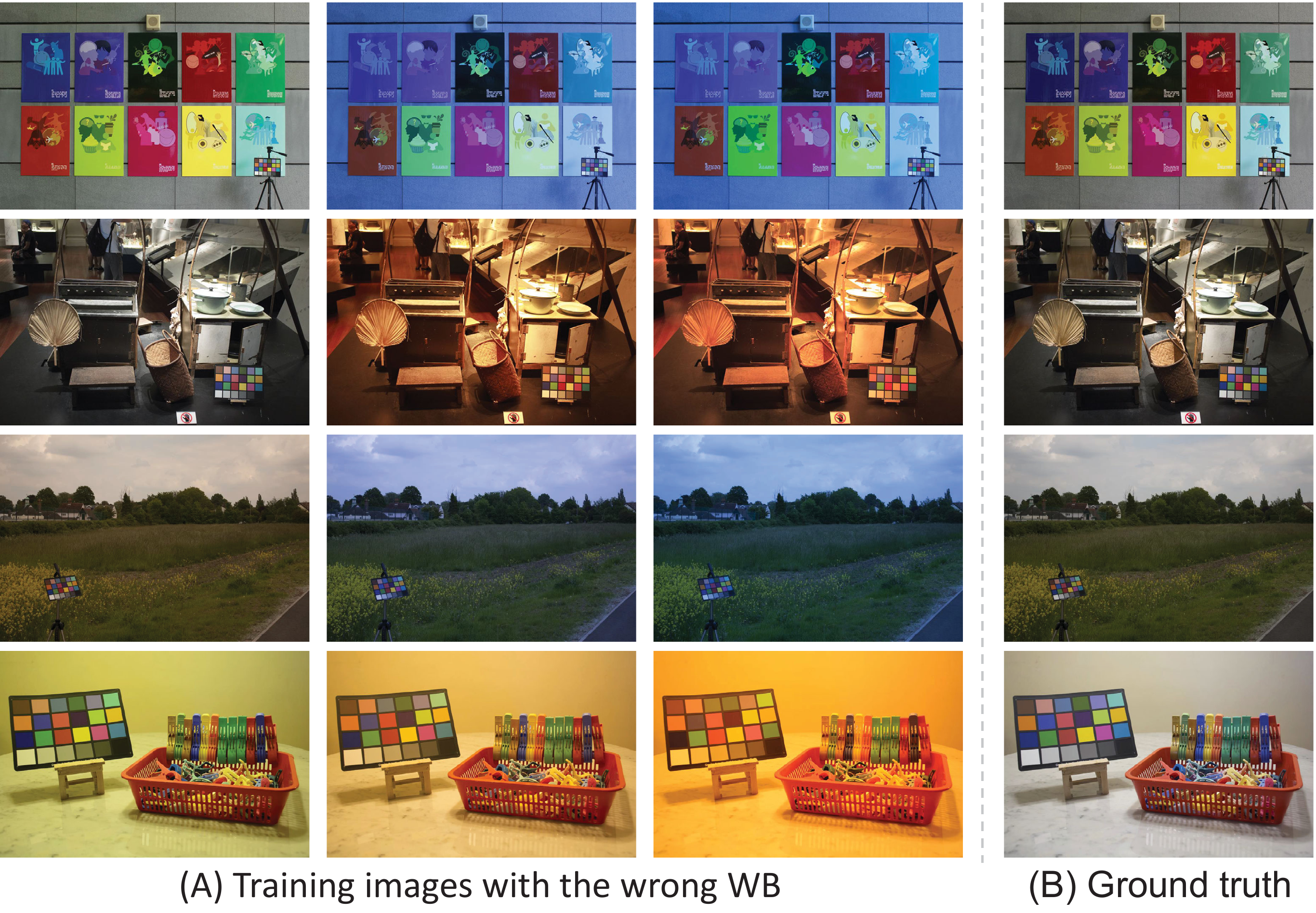}
 \caption{This figure shows examples from the Rendered WB dataset \cite{afifi2019color} used in order to generate our training rectification functions. (A) Three examples of the same scene rendered with different incorrect WB settings. (B) The ground truth white-balanced image.\vspace{-4mm}}
 \label{fig:dataset}
 \end{figure}

\subsection{3.1. Training Phase}  \vspace{1mm}

The first step of our method is to have a large number of training examples that exhibits a wide range of WB errors in camera-rendered images. We used the Rendered WB dataset proposed in \cite{afifi2019color}. This dataset contains $\sim$65,000 pairs of improperly white-balanced camera-rendered (sRGB) images and their corresponding ground truth white-balanced sRGB images. All images have WB settings applied to the original raw-RGB image followed by an emulation of in-camera nonlinear rendering operations via Adobe Lightroom~\cite{kelby2014adobe} to get the final camera-rendered image. This dataset consists of two sets: (i) the training set, referred to as Set 1, and (ii) the testing set, referred to as Set 2. Fig. \ref{fig:dataset} shows example images taken from the Rendered WB dataset~\cite{afifi2019color}.

We used all training images in Set 1 to compute the correction functions $\mathbf{M}$ described in Eq.~\ref{eq:polynomial}. For each pair, $i$, of an improperly white-balanced image, $\mathbf{I}_{\texttt{tr}(i)}$, and the corresponding ground truth image, $\mathbf{G}_{\texttt{tr}(i)}$, we compute our 33-dimensional vectorized polynomial mapping matrix $\mathbf{M}_{\texttt{tr}(i)}$ as follows:
\begin{equation}
\label{minimization_for_M}
\underset{\mathbf{M}_{\texttt{tr}(i)}}{\argmin} \left\|r\left(\mathbf{M}_{\texttt{tr}(i)}\right)\textrm{ }\phi\left(\mathbf{I}_{\texttt{tr}(i)}\right)    - \mathbf{G}_{\texttt{tr}(i)}\right\|_{\textrm{F}},
\end{equation}
\noindent
where $\left\|.\right\|_{\textrm{F}}$ is the Frobenius norm. Afterwards, each training image, $\mathbf{I}_{\texttt{tr}(i)}$, is associated with a color-cast vector  $\pmb{\gamma}^{'}_{\texttt{tr}(i)}$, which will be later used to compute a color-cast-correction vector $\pmb{\ell}_{\texttt{tr}(i)}$. This color-cast vector can be computed using any off-the-shelf illuminant estimation algorithm, $E: E\left(\mathbf{I}_{\texttt{tr}(i)}\right) \rightarrow \pmb{\gamma}^{'}_{\texttt{tr}(i)}$.

We cluster the training data based on their color-cast vectors into $k$ clusters. In our experiments, we used k-means++ \cite{arthur2006k}  with a cosine similarity distance metric and set $k$ to 50.  Each cluster, noted as $\textbf{c}$, also has a number of $\mathbf{M}_{i}$ mapping functions associated with it, where $i \in \textbf{c}$.  Instead of storing all of these mapping functions, we derive a single mapping function, termed a {\it rectification function}, that can estimate the color-mapping function based on the polynomial matrix $\mathbf{M}$. This is described in the following section.

\subsection{3.2. Rectification Function}  \vspace{1mm}

Our rectification function is inspired by a bias-correction method proposed to rectify scene illuminant estimation errors \cite{finlayson2013corrected}. Specifically, we propose a rectification function, $\mathbf{H}$, that maps a color-cast-correction vector, $\pmb{\ell}$, directly to a nonlinear correction matrix as follows:
\begin{equation}
\label{matrix_decomspition}
\mathbf{M} = \mathbf{H}\textrm{ }\pmb{\ell},
\end{equation}
\noindent
where $\mathbf{M}$ is a $33\!\times\!1$ vectorized polynomial matrix computed to map the colors of an incorrectly white-balanced image, $\mathbf{I}$, into the corresponding colors of the correctly white-balanced image, $\mathbf{G}$, $\mathbf{H}$ is our $33\!\times\!3$ rectification matrix, and $\pmb{\ell}$ is the color-cast-correction vector computed from the color-cast vector as described earlier.  This type of correction function was used by Finlayson~\cite{finlayson2013corrected} to correct biases made by illumination estimation algorithms.  In ~\cite{finlayson2013corrected}, an estimated illumination vector would be mapped to a new illumination vector that was closer to the ground truth based on the training dataset.  In our case, the function $\mathbf{H}$ maps $\pmb{\ell}$ to its corresponding matrix $\mathbf{M}$, allowing us to connect a color-cast vector directly to a mapping function.

Working from Eq.~\ref{matrix_decomspition}, we compute a rectification matrix for each cluster, denoted as $\mathbf{H}_{\textbf{c}(j)}$ for cluster $j$. Let $n$ be the number of training examples belonging to each cluster $\textbf{c}(j)$, s.t. $j\in\left[1,2...k\right]$. For each cluster, we minimize the following equation to compute its rectification matrix:
\begin{equation}
\label{minimization_for_H}
\underset{\mathbf{H}_{\textbf{c}(j)}}{\argmin} \left\|\sum_{i \in \textbf{c}(j)}\mathbf{H}_{\textbf{c}(j)}\textrm{ }\pmb{\ell}_{i} - \mathbf{M}_{i}\right\|_{\textrm{F}},
\end{equation}
\noindent
where $\pmb{\ell}_{i}$ is a $3\!\times\!1$ color-cast-correction vector of the $i^\text{th}$ training example in cluster $\textbf{c}(j)$, $\mathbf{M}_{i} \in \mathbb{R}^{33}$ is a vectorized polynomial matrix associated with the $i^\text{th}$ training example in cluster $\textbf{c}(j)$, and $\mathbf{H}_{\textbf{c}(j)}$ is a $33\!\times\!3$ rectification matrix assigned to cluster $\textbf{c}(j)$.   Eq.~\ref{minimization_for_H} essentially estimates a single $\mathbf{H}$ per cluster that minimizes the error over all $\pmb{\ell}_{i}$ associated with this cluster.

After this procedure, each color-cast cluster is now represented by the mean color-cast vector, $\pmb{\gamma}_{c_j}$, of all color-cast vectors that belong to it. For each cluster, we store the corresponding rectification matrix, $\mathbf{H}_{\textbf{c}(j)}$, to be used in the testing phase. This model requires only 0.04 MB of memory to encode the 50 rectification functions, compared to $\sim$25 MB required in~\cite{afifi2019color}.  This represents a $\sim$99\% reduction in memory requirements.

\subsection{3.3. Testing Phase}  \vspace{1mm}

Our method can easily be used in an AWB mode. To do this, given an input image $\mathbf{I}$, the same illuminant estimation algorithm, $E$, used in the training phase to compute the illuminant vector, $\pmb{\gamma}^{'}$, is applied.  Afterwards, we search among the pre-computed color-cast cluster centers to find the closest cluster $\textbf{c}(h)$ to our testing color vector. Then, the rectification function of the closest cluster is retrieved and used along with the computed color-cast-correction vector, $\pmb{\ell}^{'}$, to correct the testing image as described in the following equation:
\begin{equation}
\label{our_correction}
\mathbf{I}_{\texttt{corr}} = r\left(\mathbf{H}_{\textbf{c}(h)}\textrm{ } \pmb{\ell}^{'}\right)\textrm{ }\phi\left(\mathbf{I}\right).
\end{equation}

\begin{figure*}[t]
\centering
\includegraphics[width=0.93\linewidth]{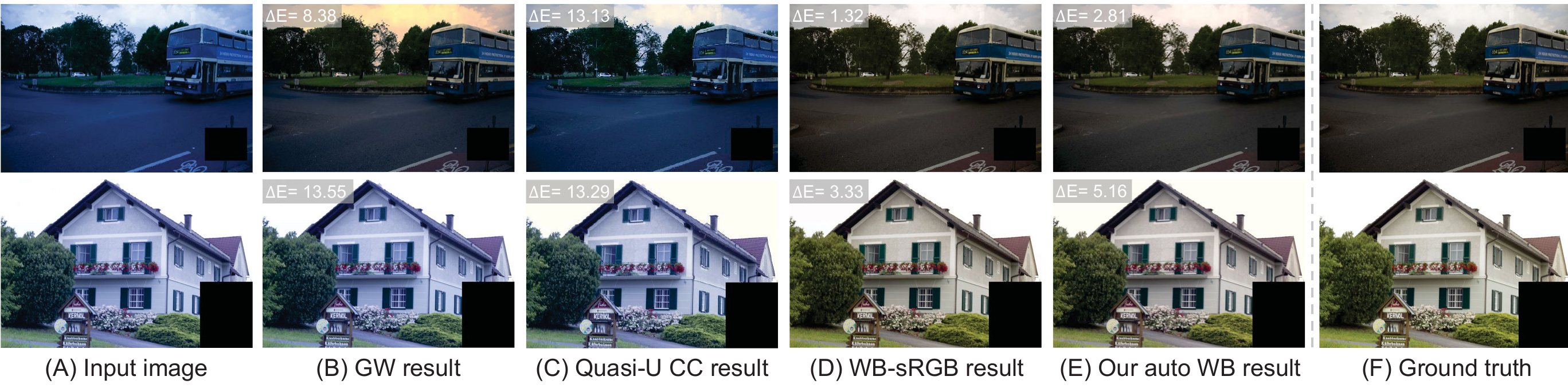}
\caption{Qualitative results on the Rendered WB dataset \cite{afifi2019color} and the rendered version of the Cube+ dataset \cite{banic2017unsupervised, afifi2019color}. (A) Input images. (B) Diagonal correction using the gray-world (GW) method \cite{buchsbaum1980spatial}. (C) Diagonal correction using the quasi-unsupervised color constancy (quasi-U CC) method \cite{bianco2019quasi}. (D) Results of the WB-sRGB method \cite{afifi2019color}. (E) Results of applying our rectification function to initial estimation of GW. (F) Ground truth images.\vspace{-2mm}}
\label{fig:qualitative_results}
\end{figure*}

\begin{figure*}[t]
\centering
\includegraphics[width=0.94\linewidth]{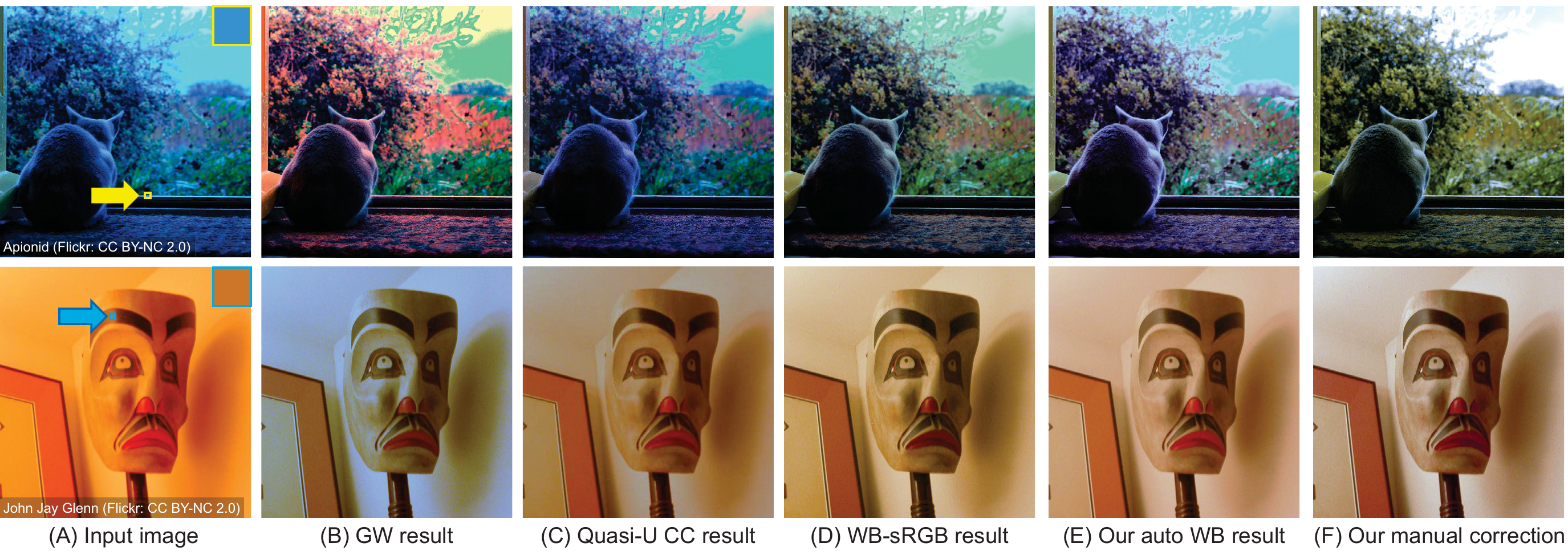}
\caption{Our method allows the user to manually select achromatic points in order to improve the results. (A) Input images. (B) Results of the gray-world (GW) method \cite{buchsbaum1980spatial}. (C) Results of the quasi-unsupervised color constancy (quasi-U CC) method \cite{bianco2019quasi}. (D) Results of the WB-sRGB method \cite{afifi2019color}. (E) Our results using the GW's estimated illuminant colors. (F) Our results using manually selected achromatic reference points.\vspace{-4mm}}
\label{fig:manual}
\end{figure*}

In the user-interactive mode, we use the color-cast vector supplied by the user of some achromatic reference point in the image instead of the estimated color vector $\pmb{\gamma}^{'}$. Our method requires $\sim$0.5 seconds to correct a 12-mega-pixel image on an Intel$^\circledR$ Xeon$^\circledR$ E5-1607 @ 3.10 GHz machine, compared to 1.5 seconds required in~\cite{afifi2019color} using a similar machine.  Our approach can significantly improve the results of diagonal-based methods (e.g., \cite{buchsbaum1980spatial, GE, SoG, hu2017fc}) to correct improperly white-balanced images.

\section{4. Experimental Results}  \vspace{3mm}
We evaluate our method extensively through quantitative and qualitative comparisons with existing solutions for AWB and user-interactive WB correction. Sec.\ 4.1 provides quantitative evaluation of our method, while qualitative comparisons are provided in Sec.\ 4.2.

 \begin{table*}[t]
 \centering
 \caption{Table 1. \hspace{1mm} Quantitative results on the Rendered WB testing set (Set 2) \cite{afifi2019color} and the rendered version of the Cube+ dataset \cite{banic2017unsupervised,afifi2019color}. We applied our rectification function (RF) to the estimated illuminants of different methods. The term ``linearized'' refers to applying the standard gamma linearization \cite{anderson1996proposal, ebner2007color} to images before estimating and correcting images. The terms Q1, Q2, and Q3 denote the first, second (median), and third quartile, respectively. The terms MSE and MAE stand for mean square error and mean angular error, respectively.  The best results are highlighted in yellow and boldface. As shown, we achieve on par results with the WB-sRGB method \cite{afifi2019color} by applying our post-processing rectification to different illuminant estimation methods. In contrast to the WB-sRGB method\cite{afifi2019color}, our method allows the user to interactively adjust the results by manually changing the selected achromatic reference point.}
 \label{Table0}
 \vspace{8mm}
 \scalebox{0.8}{
 \begin{tabular}{|l|c|c|c|c|c|c|c|c|c|c|c|c|}
 \hline
 \multicolumn{1}{|c|}{} & \multicolumn{4}{c|}{\textbf{MSE}} & \multicolumn{4}{c|}{\textbf{MAE}} & \multicolumn{4}{c|}{\textbf{$\boldsymbol{\bigtriangleup}$\textbf{E} 2000}} \\ \cline{2-13}
 \multicolumn{1}{|c|}{\multirow{-2}{*}{\textbf{Method}}} & \textbf{Mean} & \textbf{Q1} & \textbf{Q2} & \textbf{Q3} & \textbf{Mean} & \textbf{Q1} & \textbf{Q2} & \textbf{Q3} & \textbf{Mean} & \textbf{Q1} & \textbf{Q2} & \textbf{Q3}  \\ \hline

 \multicolumn{13}{|c|}{\cellcolor[HTML]{99FF66}\textbf{Testing set of the Rendered WB dataset (Set 2): DSLR and mobile phone cameras (2,881 images)}} \\ \hline
 GW \cite{buchsbaum1980spatial} & 500.18 & 173.69 & 332.75 & 615.40 & 8.89\textdegree & 5.82\textdegree & 8.32\textdegree & 11.33\textdegree & 10.74 & 7.92 & 10.29 & 13.12\\ \hline
 SoG \cite{SoG} & 429.35 & 147.05 & 286.84 & 535.72 & 9.54\textdegree & 5.72\textdegree & 8.85\textdegree & 12.65\textdegree & 10.01 & 7.09 & 9.85 & 12.69\\ \hline
 FC4 \cite{hu2017fc} & 662.53  & 304.88 & 524.42 & 817.57 & 8.92\textdegree & 5.94\textdegree & 8.03\textdegree & 10.84\textdegree & 12.12 &  8.94& 11.79 &  14.76 \\ \hdashline
 GW (linearized) \cite{buchsbaum1980spatial} & 469.86 & 163.07 & 312.28 & 574.85 & 8.61\textdegree & 5.44\textdegree & 7.94\textdegree & 10.93\textdegree & 10.68 & 7.70 & 10.13 & 13.15\\ \hline

 SoG (linearized) \cite{SoG} & 393.85 & 137.21 & 267.37 & 497.40 & 8.96\textdegree & 5.31\textdegree & 8.26\textdegree & 11.97\textdegree & 9.81 & 6.87 & 9.67 & 12.46\\ \hline

 FC4 (linearized) \cite{hu2017fc}  &  505.30 & 142.46 & 307.77 & 635.35 & 10.37\textdegree
 & 5.31\textdegree & 9.26\textdegree & 14.15\textdegree & 10.82 & 7.39 & , 10.64 &  13.77   \\ \hdashline
 \cellcolor[HTML]{CCCCCC}GW \cite{buchsbaum1980spatial} + our RF &  207.13 &  46.71 &  111.89 & 230.10 & 5.35\textdegree & 2.89\textdegree &  4.59\textdegree  &  6.84\textdegree & 6.74 & 4.45 & 6.15 & 8.45\\ \hline
 \cellcolor[HTML]{CCCCCC}SoG \cite{SoG}  + our RF & 256.10 &  55.93 & 132.09 & 266.61 & 6.25\textdegree & 3.28\textdegree &  5.20\textdegree & 8.20\textdegree & 7.27 & 4.77 & 6.61 & 9.05  \\\hline
 \cellcolor[HTML]{CCCCCC}FC4 \cite{hu2017fc} + our RF & 303.99 &  50.01 & 118.88 &  298.44 & 6.61\textdegree & 2.99\textdegree & 4.99\textdegree & 8.40\textdegree & 7.28  & 4.30 & 6.22 &  9.33 \\ \hdashline
 WB-sRGB \cite{afifi2019color} & \cellcolor[HTML]{F9FD67} \textbf{171.09} &  \cellcolor[HTML]{F9FD67} \textbf{37.04} &  \cellcolor[HTML]{F9FD67} \textbf{87.04} &  \cellcolor[HTML]{F9FD67} \textbf{190.88} &  \cellcolor[HTML]{F9FD67} \textbf{4.48\textdegree} &  \cellcolor[HTML]{F9FD67} \textbf{2.26\textdegree} &  \cellcolor[HTML]{F9FD67} \textbf{3.64\textdegree} &  \cellcolor[HTML]{F9FD67} \textbf{5.95\textdegree} &  \cellcolor[HTML]{F9FD67} \textbf{5.60} &  \cellcolor[HTML]{F9FD67} \textbf{3.43} &  \cellcolor[HTML]{F9FD67} \textbf{4.90} &  \cellcolor[HTML]{F9FD67} \textbf{7.06}  \\ \hline
 \multicolumn{13}{|c|}{\cellcolor[HTML]{99FF66}\textbf{Rendered Cube+ dataset with different WB settings (10,242 images)}} \\ \hline
 GW \cite{buchsbaum1980spatial} & 312.62 &  55.16 & 159.63 & 358.02 & 6.85\textdegree & 3.08\textdegree & 5.76\textdegree & 9.70\textdegree & 9.01 & 5.35 & 8.38 & 12.08 \\ \hline

 SoG \cite{SoG}  &  269.31 & 21.92 & 90.37 & 312.02 & 6.69\textdegree & 2.3\textdegree & 4.63\textdegree & 9.62\textdegree & 7.70 & 3.40 & 6.38 & 11.07\\ \hline
 FC4 \cite{hu2017fc} &  410.01 & 79.26 & 219.05 & 505.71 & 6.7\textdegree & 3.26\textdegree & 5.45\textdegree & 8.7\textdegree & 10.4 & 6.51 & 9.73 & 13.43 \\ \hdashline
 GW (linearized) \cite{buchsbaum1980spatial}  &  244.59 & 32.58 & 121.42 & 300.99 & 6.37\textdegree & 2.51\textdegree & 5.13\textdegree & 9.09\textdegree & 8.05 & 4.18 & 7.25 & 11.08 \\ \hline

 SoG (linearized) \cite{SoG}  & 275.33  & 17.16 & 67.49  & 309.97 & 6.66\textdegree & 2.08\textdegree & 4.17\textdegree & 9.58\textdegree & 7.57 & 3.00 & 5.73 & 11.05 \\ \hline

 FC4 (linearized) \cite{hu2017fc}  & 371.9 & 79.15 & 213.41 & 467.33 & 6.49\textdegree & 3.34\textdegree & 5.59\textdegree & 8.59\textdegree & 10.38 & 6.6 & 9.76 & 13.26 \\ \hdashline
 \cellcolor[HTML]{CCCCCC}GW \cite{buchsbaum1980spatial} + our RF  & \cellcolor[HTML]{F9FD67} \textbf{159.88} & 21.94 & 54.76 & 125.02 & 4.64\textdegree & 2.12\textdegree & 3.64\textdegree & 5.98\textdegree & 6.2 & 3.28 & 5.17 & 7.45 \\ \hline
 \cellcolor[HTML]{CCCCCC}SoG \cite{SoG} + our RF  & 226.83 & 20.01 & 58.61 & 165.03 & 5.33\textdegree & 2.1\textdegree & 3.83\textdegree & 6.97\textdegree & 6.61 & 3.17 & 5.38 & 8.56 \\ \hline
 \cellcolor[HTML]{CCCCCC}FC4 \cite{hu2017fc} + our RF & 175.73 & \cellcolor[HTML]{F9FD67} \textbf{17.8} & \cellcolor[HTML]{F9FD67} \textbf{43.65} & \cellcolor[HTML]{F9FD67} \textbf{114.65} & 4.67\textdegree & 1.89\textdegree & \cellcolor[HTML]{F9FD67} \textbf{3.10\textdegree} & 5.48\textdegree & 5.7 & \cellcolor[HTML]{F9FD67} \textbf{2.95} & 4.63 & 7.05 \\ \hdashline
 WB-sRGB \cite{afifi2019color} &  194.98 &  27.43 &  \cellcolor[HTML]{F9FD67} \textbf{57.08} &  \cellcolor[HTML]{F9FD67} \textbf{118.21} &  \cellcolor[HTML]{F9FD67} \textbf{4.12\textdegree} &  \cellcolor[HTML]{F9FD67} \textbf{1.96\textdegree} & 3.17\textdegree &  \cellcolor[HTML]{F9FD67} \textbf{5.04\textdegree} &  \cellcolor[HTML]{F9FD67} \textbf{5.68} &  3.22 &  \cellcolor[HTML]{F9FD67} \textbf{4.61} &  \cellcolor[HTML]{F9FD67} \textbf{6.70} \\\hline
 \end{tabular}
 }
 \end{table*}

 \begin{figure*}[t]
 \centering
 \includegraphics[width=0.94\linewidth]{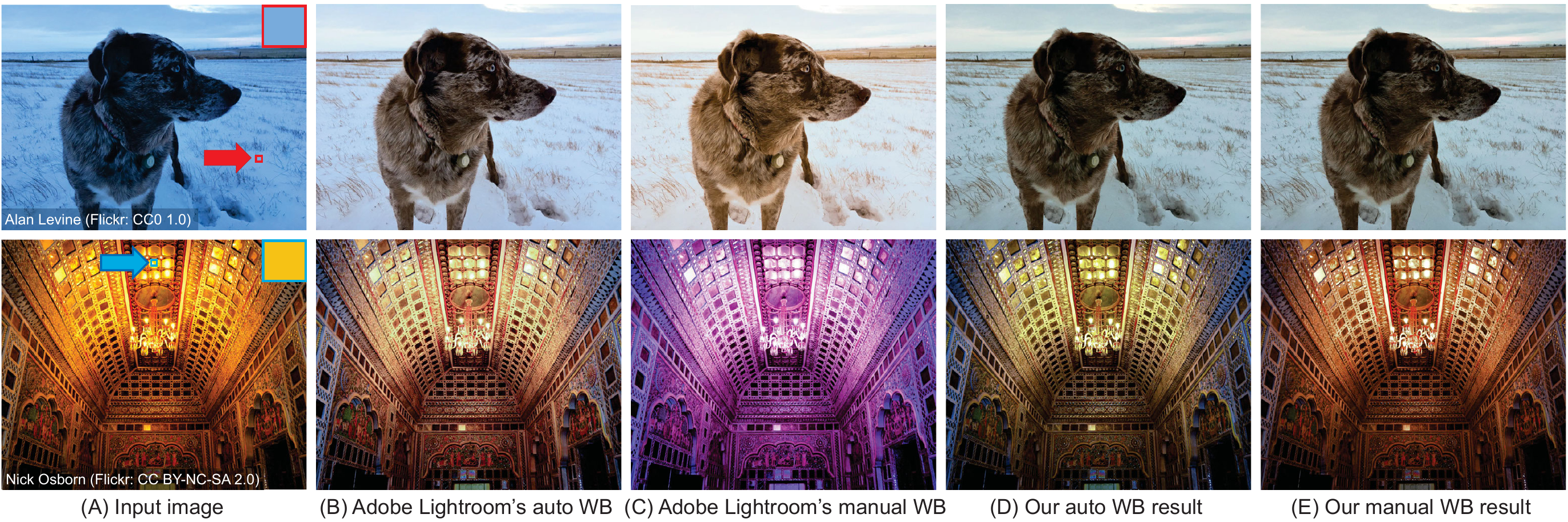}
 \caption{Comparison with Adobe Lightroom WB correction. (A) Input image. (B) and (C) Adobe Lightroom's auto WB and manual WB correction results, respectively. (D) and (E) Our auto and manual WB correction results, respectively.\vspace{-4mm}}
 \label{fig:lightroom}
 \end{figure*}

\subsection{4.1. Quantitative Evaluation}  \vspace{1mm}

We evaluated our AWB correction results using $\sim$13,000 testing images from the Rendered WB testing set (Set 2) \cite{afifi2019color} and the rendered version of the Cube+ dataset \cite{banic2017unsupervised,afifi2019color}.  As mentioned in Sec. 3, any off-the-shelf illumination estimation method $E$ can be used. We tested different illuminant estimation methods in our AWB framework.  Specifically, we utilized the following methods: the gray-world (GW) \cite{buchsbaum1980spatial}, the shades-of-gray (SoG) \cite{SoG}, and the FC4 \cite{hu2017fc} methods. In each experiment, we use the training data of the Rendered WB dataset \cite{afifi2019color} to compute our rectification functions, as described in Sec.\ 3.2. For each of the illuminant estimation methods, we compare our results with the diagonal correction with and without linearizing the testing images. The linearizing process was performed using the standard de-gamma linearization operation \cite{anderson1996proposal, ebner2007color}. We include this linearization process in our comparisons as it is a common misconception that a simple gamma linearization can remove the nonlinearity applied by cameras.

For the sake of completeness, we also compare our results with the recent nonlinear method for post-capture WB correction \cite{afifi2019color}. In this section, we refer to Afifi et al.'s method as the {\it WB-sRGB} method. Table \ref{Table0} shows the first, second, and third quantile and the mean of the error values obtained by each method. We followed the evaluation metrics used in \cite{afifi2019color}, which are: (i) mean square error (MSE), (ii) mean angular error (MAE), and (iii) $\boldsymbol{\bigtriangleup}$\textbf{E} 2000 \cite{sharma2005ciede2000}.

As shown in Table \ref{Table0}, our rectification function significantly improves the results of diagonal-based methods and achieves results on a par with the WB-sRGB method on both testing sets. As can be seen from the results, our method reduces the MSE by $\sim$55\%, the MAE by $\sim$32\%, and the $\bigtriangleup$E 2000 by $\sim$35 \% on average compared to the gamma linearization process, which reduces the MSE by $\sim$15\%, the MAE by $\sim$5\%, and the $\bigtriangleup$E 2000 by $\sim$5\% on average.

\subsection{4.2. Qualitative Evaluation}  \vspace{1mm}
We qualitatively evaluated our method against different methods, including a commercial photo-editing software, for auto and user-interactive WB correction.\ Fig.\ \ref{fig:qualitative_results} shows a comparison between our results and the diagonal correction of two illuminant estimation methods---namely, the GW method \cite{buchsbaum1980spatial} and the quasi-unsupervised color constancy method \cite{bianco2019quasi}. As shown, our AWB is superior to diagonal WB and achieves similar results to the recent WB-sRGB method \cite{afifi2019color}.

As previously mentioned, in contrast to the WB-sRGB method \cite{afifi2019color}, our method allows interactive correction to improve the results by manually adjusting the color-cast color. Fig.\ \ref{fig:manual} shows that this manual adjustment feature produces arguably visually superior results compared to the WB-sRGB method \cite{afifi2019color}.

We further compared our method against Adobe Lightroom, as it is one of the most common photo-editing software programs that provide the same manual WB correction feature. As can be seen in Fig. \ref{fig:lightroom}, our method produces perceptibly superior results in comparison with Adobe Lightroom's results for both auto and manual WB correction.

\section{5. Conclusion}  \vspace{1mm}
We have introduced an interactive WB method for use on camera-rendered images that allows the user to directly specify scene points to be used for white balancing.  Previously, this type of interaction could be performed only by photo-editing software operating on RAW images.   We have enabled this feature for camera-rendered images that already have a white-balance correction applied as well as additional photo-finishing.  Our method works by efficiently computing nonlinear color-correction mappings based on user-supplied color-cast vectors directly from the camera-rendered image.  We also showed how our method can easily perform auto-WB correction in a camera-rendered image.  Our method enables a new photo-editing feature for color manipulation.

\section*{Acknowledgements}
This study was funded in part by the Canada First Research Excellence Fund for the Vision: Science to Applications (VISTA) programme and an NSERC Discovery Grant.

{

}
\end{document}